\documentclass{article}
\usepackage{spconf,amsmath,graphicx,hyperref}

\usepackage{amsmath}
\usepackage{amsfonts}
\usepackage{booktabs}

\title{Align to the Pivot: Dual Alignment with Self-Feedback for Multilingual Math Reasoning}
\name{Chunxu Zhao\textsuperscript{1}$^{*}$, 
      Xin Huang\textsuperscript{1}$^{*}$,
      Xue Han\textsuperscript{1}$^{\dagger}$,
      Shujian Huang\textsuperscript{2},
      Chao Deng\textsuperscript{1},
      Junlan Feng\textsuperscript{1}$^{\dagger}$}
\address{\textsuperscript{1}JIUTIAN Research, China Mobile, Beijing, China\\\textsuperscript{2}National Key Laboratory for Novel Software Technology, Nanjing University}
\begin{document}
\ninept
\maketitle
\begin{abstract}
Despite the impressive reasoning abilities demonstrated by large language models (LLMs), empirical evidence indicates that they are not language agnostic as expected, leading to performance declines in multilingual settings, especially for low-resource languages. We attribute the decline to the model's inconsistent multilingual understanding and reasoning alignment. To address this, we present Pivot-Aligned Self-Feedback Multilingual Reasoning (PASMR), aiming to improve the alignment of multilingual math reasoning abilities in LLMs. This approach designates the model's primary language as the pivot language. During training, the model first translates questions into the pivot language to facilitate better alignment of reasoning patterns. The reasoning process in the target language is then supervised by the pivot language's reasoning answers, thereby establishing a cross-lingual self-feedback mechanism without relying on external correct answers or reward models. Extensive experimental results demonstrate that our method enhances both the model's understanding of questions and its reasoning capabilities, leading to notable task improvements. \footnote{Code, Data and Appendix: https://github.com/Rover912/PASMR}
\end{abstract}
\section{Introduction}
\label{sec:intro}

\footnotetext[2]{*\,Equal contribution \quad $\dagger$\,Corresponding authors}

Large language models have attracted widespread attention for their advanced reasoning capabilities \cite{jaech2024openai}. Although it may seem intuitive that reasoning abilities are language-agnostic, empirical evidence reveals that these models perform differently across languages, with significant performance drops in low-resource languages \cite{shi2022language}. Therefore, enhancing the consistency of reasoning capabilities across languages remains a critical research challenge.

\begin{figure}[t]
\centering 
\includegraphics[scale=0.48]{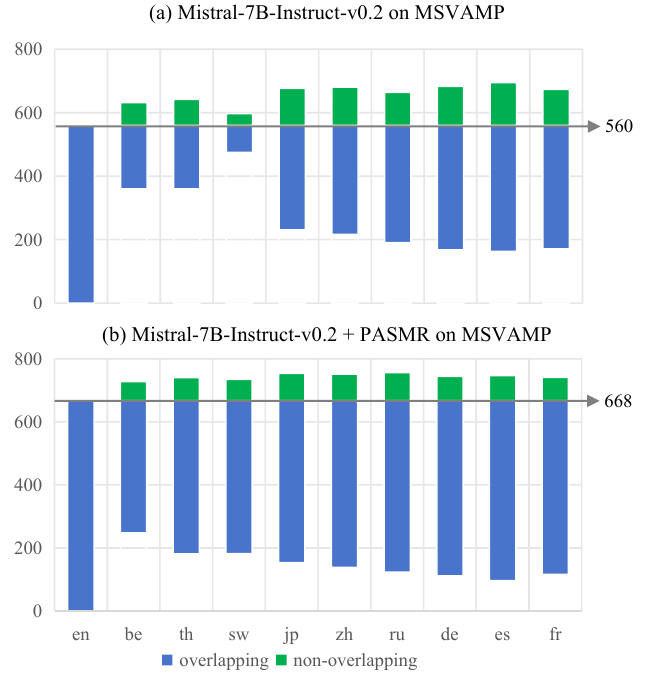} 
\caption{Number of correct answers overlapping with English across 10 languages. Blue bars: overlapped answers; green bars: non-overlapped. (a) Mistral-7B-Instruct on MSVAMP; (b) after applying our PASMR method.} 
\label{match_to_en} 
\end{figure}

Many studies \cite{wendler2024llamas,zhao2024large} show that LLMs process multilingual tasks in three steps: (1) mapping the input to a dominant language, (2) reasoning in the dominant language, and (3) mapping  back to the target language. As Figure \ref{match_to_en}(a) illustrates, our testing shows non-English answers align with English, but with lower accuracy and incomplete matches. We argue that these gaps arise from difficulties in both understanding multilingual inputs and mapping reasoning results back into the target language.

To address these issues, we propose a novel training framework, Pivot-Aligned Self-Feedback Multilingual Reasoning (PASMR), which aims to reduce the performance degradation in multilingual math reasoning tasks. The PASMR framework consists of two stages: Pivot-Aligned Mapping (PAM) and Self-feedback Reinforcement Learning (SRL). In the PAM stage, the dominant language of a LLM is chosen as the pivot language.
Then pivot questions are derived from translating multilingual questions into the pivot language to facilitate better alignment of reasoning patterns.
This alignment helps to reduce the bias introduced during the initial mapping of the problem to the pivot language. In the SRL stage, the model leverages self-generated feedback signals to reinforce the consistency of the reasoning process without relying on external correct answers or reward models. This self-feedback mechanism allows the model to iteratively improve its reasoning capabilities through online reinforcement learning, thereby enhancing its overall performance in multilingual reasoning. Our experimental results show that the PASMR framework effectively improves the multilingual reasoning capabilities of LLMs, particularly in low-resource languages. The results shown in Figure \ref{match_to_en} (b) clearly illustrate the effectiveness of PASMR. It enhances the overall reasoning ability of the model while improving multilingual reasoning consistency.
The contributions of this paper are as follows:
\begin{itemize}
    \item We propose the PASMR framework, which enhances the model's understanding of multilingual questions and refines the reasoning process through a self-feedback mechanism.
    \item Extensive experiments show our method effectively improves LLMs’ ability to solve multilingual math problems, especially for low-resource languages, while validating its generalizability and robustness.
    \item Our analysis of cross-lingual reasoning highlights persistent challenges in multilingual comprehension and reasoning.

\end{itemize}

\begin{figure*}[t] 
\centering 
\includegraphics[scale=0.45]{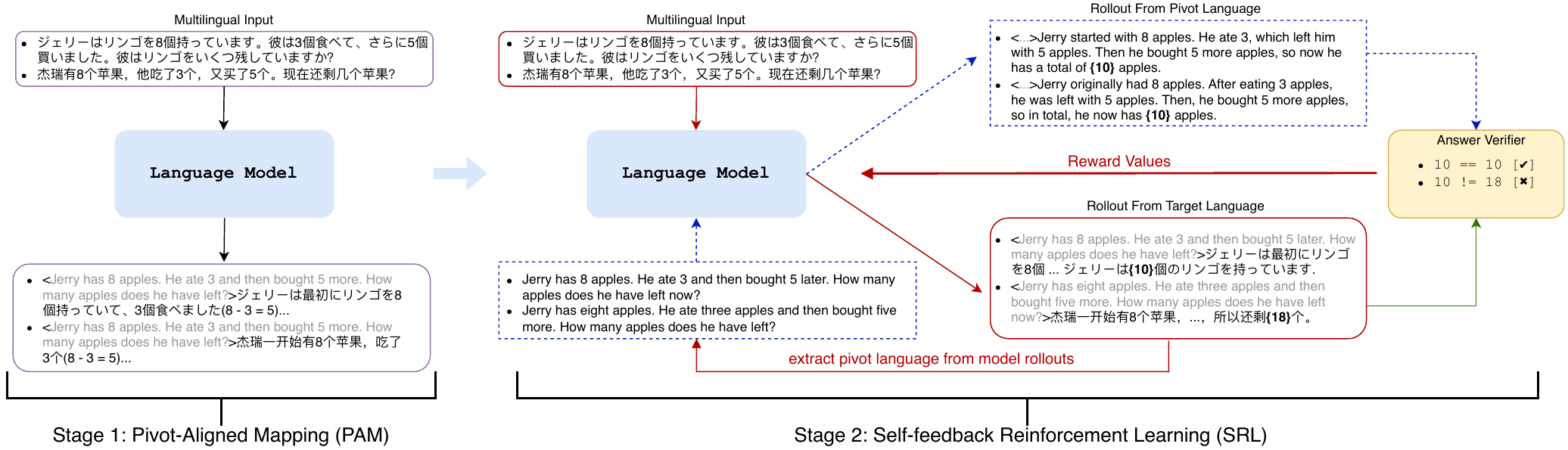}
\caption{During the PAM phase, the model is fine-tuned to map multilingual inputs to English and generate responses in the target language. In SRL, the model generates pivot rollouts after extracting pivot language from target language rollouts. A reward is then calculated by evaluating the consistency between the target language answer and the pivot language answer, which is then used to optimize the model.} 
\label{model pic} 
\end{figure*}

\section{Method}

\subsection{Pivot-Aligned Mapping}
To effectively address multilingual reasoning problems, it is crucial to improve consistent mappings of the problems across languages. Therefore, when encountering a question \( Q_T \) in the target language \( T \), we first map the question to a pivot language (English in this case) and then generate the answer in the target language. 

To achieve this, we propose an explicit mapping approach: translating multilingual questions \( Q_T \) into English \( Q_{\text{En}} \) and subsequently generating answers \( A_T \) in the target language. Specifically, given model parameters \( \theta \), the loss function is defined as:

\[
\mathcal{L}_{\text{SFT}}(\theta) = -\sum_{j=1}^N \log P_\theta(Q_{\text{En}}^{(j)}, A_T^{(j)} \mid Q_T^{(j)}),
\]

where \( Q_T^{(j)} \), \( Q_{\text{En}}^{(j)} \), and \( A_T^{(j)} \) represent the \( j \)-th sample's original language question, its English translation, and the target language answer, respectively. Note that special tokens are inserted at the beginning and end of \( Q_{\text{En}}^{(j)} \) as delimiters to facilitate extraction in the second stage, as illustrated in the bottom-left examples of Figure \ref{model pic}.

\subsection{Self-feedback Reinforcement Learning}
In multilingual reasoning, accuracy in the pivot language serves as an upper bound, since reasoning results should ideally be identical across languages but sometimes are not. To narrow this gap, we propose an online reinforcement learning method that uses the model’s own outputs as rewards to enforce consistency with the pivot language. As shown in Figure~\ref{model pic}, the model (from the PAM stage) first maps the target-language questions into English, then generates answers in the target language, and finally obtains English answers from the extracted questions. The rewards are then calculated based on the consistency between the two sets of answers and are used to optimize the model with REINFORCE++ \cite{hu2025reinforcesimpleefficientapproach}.

Formally, given a model $\pi_\theta$ and input $Q_T$, we sample an output trajectory 
$\tau = [Q_{\text{En}}, A_T]$ of length $L$, where $Q_{\text{En}}$ is the English translation of $Q_T$ 
and $A_T$ is the answer in target language $T$. Feeding $Q_{\text{En}}$ into the model yields the 
English answer $A_{\text{En}}$. The token-wise reward is

\[
R_i = 
\begin{cases}
F_{\text{reward}}(A_T, A_{\text{En}}) & i = L, \\
F_{\text{reward}}(A_T, A_{\text{En}}) - \beta \, \text{KL}(i) & i < L,
\end{cases}
\]

where $F_{\text{reward}}(A_T, A_{\text{En}}) = 1$ if $A_T = A_{\text{En}}$, $0.1$ if $A_T \neq A_{\text{En}}$, 
and $0$ if there is a format error. We set $\beta = 0.01$. $\text{KL}(i)$ is the KL divergence at token $i$, 
computed as $\text{KL}(i) = \log \frac{\pi_\theta(a_i | a_{<i})}{\pi_{\text{PAM}}(a_i | a_{<i})}$, 
where $a_i$ is the token generated at step $i$ and $a_{<i}$ is the sequence of previous tokens.

The discounted cumulative reward (advantage) for token $i$ is  

\[
G_i = \sum_{k=i+1}^{L} \gamma^{k-i} R_k, \quad \gamma=0.99,
\]

where $R_k$ is the immediate reward at step $k$. Finally, $\pi_\theta$ is optimized via  

\begin{equation*}
\begin{split}
L^{CLIP}(\theta) = \mathbb{E}_{\tau \sim \pi_\theta} \bigg[ & \sum_{i=1}^{L} \min \big( r_i(\theta) \hat{G}_i, \\
& \quad \text{clip}(r_i(\theta), 1-\epsilon, 1+\epsilon) \hat{G}_i \big) \bigg]
\end{split}
\end{equation*}

where $r_i(\theta) = \frac{\pi_\theta(a_i|a_{<i})}{\pi_{\theta_{\text{old}}}(a_i|a_{<i})}$ 
is the policy ratio, $\hat{G}_i$ is the normalized advantage, and $\epsilon = 0.2$.

\section{Experiments}
\label{sec:majhead}

\subsection{Dataset}
We construct our dataset from GSM8K \cite{cobbe2021training}, generating solutions with Qwen2.5-instruct-7B and translating both problems and solutions into nine languages (Bengali, Thai, Swahili, Japanese, Chinese, Russian, German, Spanish, French) using Qwen2.5-instruct-32B. After rule-based filtering that checks answer correctness, language consistency, and formatting, we obtain 2,048 instances per language, yielding 20,480 examples across 10 languages (including English). Construction details are provided in the supplementary materials of our repository. The dataset is evenly split between fine-tuning and reinforcement learning. To evaluate SRL robustness, we further construct an OOD train set from tasks 1, 4, and 8 of NumGLUE \cite{mishra-etal-2022-numglue, she2024mapo}, totaling 18,240 examples.

\subsection{Evaluation \& Baselines}
We evaluate the model's performance using the MGSM \cite{shi2022language} and MSVAMP datasets \cite{chen2024breakinglanguagebarriersmultilingual}. The MGSM dataset is constructed by sampling from the GSM8K test set and translating it into the target language, serving to assess the model's in-domain performance. The second dataset, MSVAMP, extends the SVAMP dataset by translating it into multiple languages. We use it to evaluate the model's out-of-domain performance.

We compare against two types of baselines. Non-training methods include:
(1) Pipeline \cite{etxaniz2023multilingual}, which translates questions to English, answers them, then translates responses back.
(2) MCOT \cite{shi2022language}, which prompts the model with two same-language exemplars.

Training-based methods include:
(1) Multilingual Supervised Fine-Tuning (MSFT) \cite{chen2024breakinglanguagebarriersmultilingual}, which fine-tunes the model to generate CoT reasoning in multiple languages.
(2) MSFT w. Gold Answer RL \cite{luong2024reftreasoningreinforcedfinetuning}, which further applies reinforcement learning using reward signals based on gold-standard answers. (3) MAPO \cite{she2024mapo}, which leverages external translation models to generate partial preference pairs for DPO-based optimization. Implementation details (e.g., hyperparameter settings, prompts) and the reproduced QA-align \cite{zhu2024question} results, which exhibit considerable fluctuations in our experiments, are provided in the supplementary materials of our repository.

\begin{table*}[t]
\centering
\normalsize 
\renewcommand{\arraystretch}{0.6}
\resizebox{1.0\textwidth}{!}{
\begin{tabular}{l|cccccccccccc|cccccccccccc}
\toprule
 & \multicolumn{12}{c|}{\textbf{MGSM}} & \multicolumn{12}{c}{\textbf{MSVAMP}} \\
\cmidrule(lr){2-13} \cmidrule(lr){14-25}
\textbf{Method} & Bn & Th & Sw & Ja & Zh & Ru & De & Es & Fr & En & low-res. & Avg. 
                & Bn & Th & Sw & Ja & Zh & Ru & De & Es & Fr & En & low-res. & Avg. \\

\midrule

\textbf{Mistral-7B-Instruct} & 16.0 & 12.0 & 6.8 & 20.4 & 32.8 & 35.2 & 33.2 & 37.6 & 34.4 & 41.2 & 11.6 & 27.0
                             & 27.2 & 28.2 & 12.2 & 44.6 & 46.4 & 47.3 & 51.5 & 53.2 & 50.3 & 56.0 & 22.5 & 41.7 \\
\quad + Pipeline             & 18.8 & 16.0 & 3.8 &  27.0 & 29.0 & 35.2 & 36.8 & 35.6 & 33.4 & 41.6 & 12.8 & 27.7
                             & 31.1 & 31.6 & 9.6 &  49.4 & 50.0 & 50.9 & 52.1 & 52.5 & 55.0 & 56.4 & 24.1 & 43.9 \\
\quad + MCOT                 & 19.6 & 24.8 & 7.6 &  26.8 & 34.0 & 39.2 & 36.4 & 41.6 & 32.4 & 48.0 & 17.3 & 31.0
                             & 34.0 & 39.6 & 15.1 & 46.0 & 49.0 & 55.3 & 53.0 & 56.7 & 52.4 & 58.1 & 29.5 & 45.9 \\
\quad + MSFT                 & 32.0 & 41.2 & 31.2 & 35.2 & 42.0 & 42.8 & 44.0 & 49.2 & 45.6 & 52.8 & 34.8 & 41.6
                             & 36.9 & 45.5 & 43.7 & 46.3 & 48.1 & 50.0 & 49.8 & 51.3 & 50.9 & 56.4 & 42.0 & 47.9 \\
\quad+ MSFT w. Gold Answer RL & 36.8 & 44.0 & 40.4 & 43.2 & 48.8 & 53.6 & 50.0 & 56.4 & 52.8 & 61.6 & 40.4 & 48.8
                             & 41.1 & 49.7 & 45.9 & 51.1 & 54.5 & 54.3 & 53.4 & 56.6 & 56.8 & 58.9 & 45.6 & 52.2 \\
\quad + MAPO                 & 23.6 & 25.2 & 23.2 & 29.2 & 31.2 & 34.0 & 31.2 & 35.6 & 35.6 & 39.2 & 24.0 & 30.8 & 37.8 & 46.1 & 41.4 & 43.4 & 44.5 & 45.5 & 46.2 & 49.2 & 44.6 & 53.4 & 41.7 & 45.2 \\
\quad + PASMR (Ours)                & \textbf{42.4} & \textbf{45.6} & \textbf{45.2} & \textbf{46.4} & 52.8 & \textbf{54.4} & \textbf{57.6} & \textbf{60.8} & \textbf{56.4} & \textbf{63.2} & \textbf{44.4} & \textbf{52.5}
                             & \textbf{48.0} & \textbf{56.0} & 55.3 & \textbf{60.1} & 61.4 & 63.4 & 63.4 & \textbf{65.1} & 62.6 & 66.8 & \textbf{53.1}  & 60.2 \\
\quad + PASMR w. OOD data (Ours)    & 39.2 & 44.4 & 37.2 & 40.8 & \textbf{53.2} & 53.2 & 51.6 & 55.2 & 50.0 & 56.0 & 40.3 & 48.1
                             & 46.3 & 55.5 & \textbf{56.6} & 58.9 & \textbf{62.4} & \textbf{64.0} & \textbf{63.6} & 64.1 & \textbf{64.8} & \textbf{68.2} & 52.8  & \textbf{60.4} \\
\midrule

\textbf{Llama-3-8B-Instruct} & 48.4 & 55.6 & 32.0 & 49.6 & 56.8 & 64.8 & 61.6 & 67.2 & 63.2 & 78.4 & 45.3 & 57.8
                             & 60.5 & 65.4 & 60.0 & 68.9 & 70.4 & 73.1 & 70.0 & 75.8 & 72.5 & 78.2 & 62.0 & 69.5 \\
\quad + Pipeline             & 58.6 & 60.6 & 42.6 & 61.6 & 61.2 & 70.8 & 69.4 & 73.2 & 69.2 & 78.0 & 53.9 & 64.5
                             & 63.6 & 67.5 & 60.0 & 69.5 & \textbf{76.8} & \textbf{77.0} & \textbf{78.5} & \textbf{78.6} & 77.8 & 77.8 & 63.7 & 72.1 \\
\quad + MCOT                 & 49.6 & 55.6 & 44.4 & 53.6 & 56.8 & 65.2 & 66.4 & 62.8 & 54.8 & 70.8 & 49.8 & 58.0
                             & 58.9 & 67.2 & 63.9 & 69.4 & 73.0 & 71.3 & 75.6 & 72.7 & 73.6 & 77.7 & 63.3 & 69.5 \\
\quad + MSFT                 & 51.6 & 57.6 & 46.4 & 49.6 & 59.2 & 63.2 & 62.4 & 66.4 & 62.8 & 71.2 & 51.9 & 59.0
                             & 48.9 & 56.7 & 56.3 & 62.5 & 64.7 & 67.3 & 68.6 & 70.1 & 69.7 & 76.1 & 54.0 & 64.1 \\
\quad+ MSFT w. Gold Answer RL & 55.2 & 68.0 & 52.0 & 58.4 & 64.8 & 70.0 & 70.4 & 73.2 & 68.4 & 79.6 & 58.4 & 66.0
                             & 52.6 & 61.4 & 59.1 & 65.3 & 68.1 & 68.1 & 70.5 & 70.4 & 71.7 & 75.2 & 57.7 & 66.2 \\
\quad + MAPO                 & 49.2 & 60.4 & 46.4 & 50.0 & 61.6 & 62.0 & 63.6 & 68.4 & 65.2 & 74.8 & 52.0 & 60.2 & 61.3 & 66.1 & 63.3 & 70.7 & 71.1 & 73.8 & \textbf{78.0} & 78.2 & \textbf{78.0} & \textbf{82.0} & 63.5 & 72.2 \\
\quad + PASMR (Ours)                 & 59.6 & 59.6 & \textbf{57.2} & 57.6 & \textbf{66.8} & \textbf{72.8} & \textbf{73.2} & \textbf{74.4} & 67.2 & \textbf{80.4} & 58.8 & 66.9 & 61.7 & \textbf{69.2} & 66.4 & 72.7 & 69.3 & 71.0 & 75.3 & 76.1 & 74.0 & 77.8 & 65.8 & 71.3 \\
\quad + PASMR w. OOD data (Ours)    & \textbf{60.0} & \textbf{64.0} & 55.6 & \textbf{59.6} & 66.0 & 70.8 & 72.4 & 73.2 & \textbf{68.8} & 79.6 & \textbf{59.9} & \textbf{67.0}
                             & \textbf{65.7} & 69.1 & \textbf{69.5} & \textbf{74.7} & 74.1 & 73.7 & 78.0 & 78.4 & 77.1 & 77.6 & \textbf{68.1}  & \textbf{74.0} \\
\midrule

\textbf{Deepseek-math-7b-instruct} & 46.4 & 58.0 & 13.2 & 64.0 & 72.8 & 72.4 & 69.2 & 70.0 & 67.6 & 84.0 & 39.2 & 61.8
                             & 58.1 & 70.1 & 26.0 & 78.0 & 82.0 & 75.4 & 77.2 & 78.0 & 75.2 & 80.0 & 51.4 & 70.0 \\
\quad + Pipeline             & 16.6 & 29.6 & 5.6 &  32.6 & 25.2 & 51.8 & 44.8 & 45.6 & 49.8 & 37.6 & 17.2 & 33.9
                             & 28.5 & 36.8 & 13.8 & 37.7 & 21.9 & 56.6 & 46.0 & 43.9 & 52.4 & 43.8 & 26.3 & 38.1 \\
\quad + MCOT                 & 46.0 & 56.4 & 13.6 & 63.2 & 71.6 & 72.4 & 65.2 & 69.2 & 64.4 & 78.4 & 38.6 & 60.0
                             & 54.6 & 65.9 & 35.8 & 77.6 & 79.8 & 76.5 & 76.6 & 78.6 & 75.8 & 77.4 & 52.1 & 69.9 \\
\quad + MSFT                 & 56.0 & 65.6 & 40.4 & 68.0 & 72.4 & 76.0 & 74.8 & 77.2 & 75.6 & 86.0 & 54.0 & 69.2
                             & 58.1 & 69.2 & 52.8 & 73.0 & 76.5 & 77.9 & 79.6 & 81.1 & 79.4 & 84.1 & 60.0 & 73.2 \\
\quad+ MSFT w. Gold Answer RL & 59.2 & \textbf{70.4} & 44.0 & \textbf{73.6} & 74.0 & \textbf{80.4} & 78.8 & 78.0 & 73.6 & 85.2 & 57.9 & 71.7
                             & 62.3 & 70.2 & 54.0 & 77.7 & 80.9 & 81.4 & 80.5 & 82.4 & 79.5 & 83.6 & 62.2  & 75.2 \\
\quad + MAPO                 & \textbf{62.8} & 67.6 & 44.0 & 70.0 & 75.2 & 77.6 & 77.2 & 78.8 & 74.8 & 85.2 & 58.1 & 71.3 & 67.4 & 71.8 & 61.1 & 79.0 & 82.0 & 82.0 & 81.8 & \textbf{84.1} & 82.5 & 85.6 & 66.7 & 77.7 \\
\quad + PASMR (Ours)                & 60.8 & 69.6 & \textbf{49.2} & 69.2 & \textbf{76.4} & 77.2 & \textbf{78.8}& 79.6 & \textbf{76.4} & \textbf{86.4} & \textbf{59.9} & \textbf{72.4}
                             & \textbf{68.2} & \textbf{75.1} & \textbf{69.7} & \textbf{80.9} & \textbf{82.4} & 82.0 & 81.2 & 83.1 & 82.6 & 85.4 & \textbf{71.0}  & \textbf{79.1}\\
\quad + PASMR w. OOD data (Ours)    & 60.4 & 68.4 & 47.2 & 66.8 & \textbf{76.4} & 77.6 & 78.4 & \textbf{80.4} & 76.0 & 84.8 & 58.7 & 71.6
                             & 66.1 & 74.7 & 66.5 & 79.4 & 82.1 & \textbf{82.7} & \textbf{82.5} & 83.9 & \textbf{84.1} & \textbf{86.9} & 69.1 & 78.9\\

\bottomrule
\end{tabular}
}
\caption{Results on MGSM and MSVAMP. Bold marks the best within the same base models. ``low-res.'' is the average of Bn, Th, and Sw.}
\label{table:merged_results}
\end{table*}

\begin{table*}[t]
\centering
\normalsize 
\renewcommand{\arraystretch}{0.6}
\resizebox{1.0\textwidth}{!}{
\begin{tabular}{l|cccccccccccc|cccccccccccc}
\toprule
 & \multicolumn{12}{c|}{\textbf{MGSM}} & \multicolumn{12}{c}{\textbf{MSVAMP}} \\
\cmidrule(lr){2-13} \cmidrule(lr){14-25}
\textbf{Method} & Bn & Th & Sw & Ja & Zh & Ru & De & Es & Fr & En & low-res. & Avg. 
                & Bn & Th & Sw & Ja & Zh & Ru & De & Es & Fr & En & low-res. & Avg. \\
\midrule

\textbf{PASMR Mistral-7B-Instruct} & 42.4 & 45.6 & 45.2 & 46.4 & 52.8 & 54.4 & 57.6 & 60.8 & 56.4 & 63.2 & 44.4 & 52.5 & 48.0 & 56.0 & 55.3 & 60.1 & 61.4 & 63.4 & 63.4 & 65.1 & 62.6 & 66.8 & 53.1 & 60.2 \\
\textbf{wo. SRL} & 36.4 & 41.6 & 36.0 & 42.8 & 49.6 & 52.8 & 49.2 & 52.4 & 50.8 & 53.6 & 38.0 & 46.5 & 41.0 & 49.6 & 49.7 & 53.0 & 54.3 & 56.1 & 57.1 & 58.1 & 57.6 & 61.9 & 46.8 & 53.8 \\
\textbf{w. gold answer RL} & 42.4 & 47.6 & 46.8 & 44.8 & 54.8 & 54.0 & 56.4 & 59.6 & 52.8 & 64.4 & 45.6 & 52.4 & 42.5 & 49.6 & 50.6 & 57.3 & 57.7 & 60.2 & 60.6 & 63.1 & 60.3 & 62.9 & 47.6 & 56.5 \\
\textbf{Infer w. nllb pivot (PASMR)} & 46.8 & 36.8 & 46.0 & 47.2 & 47.2 & 52.0 & 57.6 & 57.6 & 54.4 & 63.6 & 43.2 & 50.9 & 48.9 & 52.6 & 54.4 & 57.5 & 60.3 & 58.5 & 61.6 & 63.4 & 61.7 & 65.5 & 52.0 & 58.4 \\
\textbf{Infer w. nllb pivot (PASMR wo. SRL)} & 39.2 & 34.8 & 36.8 & 43.2 & 45.6 & 47.2 & 50.8 & 52.0 & 51.6 & 54.4 & 36.9 & 45.6 & 41.5 & 46.4 & 50.4 & 51.7 & 53.1 & 53.4 & 56.1 & 59.2 & 53.7 & 61.3 & 46.1 & 52.7 \\
\textbf{Infer w. gold pivot  (PASMR)} & 55.2 & 58.8 & 53.2 & 56.8 & 58.8 & 60.8 & 59.6 & 60.4 & 60.0 & 63.6 & 55.7 & 58.7 & 55.7 & 61.4 & 59.3 & 60.4 & 60.6 & 62.5 & 63.9 & 64.1 & 62.6 & 65.3 & 58.8 & 61.6 \\
\textbf{Infer w. gold pivot (PASMR wo. SRL)} & 47.6 & 51.2 & 49.2 & 47.2 & 57.2 & 55.2 & 52.8 & 53.6 & 48.0 & 54.0 & 49.3 & 51.6 & 48.4 & 54.6 & 52.9 & 54.5 & 55.7 & 57.8 & 58.2 & 58.6 & 56.1 & 60.9 & 52.0 & 55.8 \\

\midrule

\textbf{PASMR Llama-3-8B-Instruct} & 59.6 & 59.6 & 57.2 & 57.6 & 66.8 & 72.8 & 73.2 & 74.4 & 67.2 & 80.4 & 58.8 & 66.9 & 61.7 & 69.2 & 66.4 & 72.7 & 69.3 & 71.0 & 75.3 & 76.1 & 74.0 & 77.8 & 65.8 & 71.3 \\
\textbf{wo. SRL} & 53.6 & 61.2 & 54.0 & 57.2 & 64.0 & 70.8 & 70.8 & 71.2 & 66.0 & 78.0 & 56.3 & 64.7 & 60.0 & 66.0 & 64.4 & 70.9 & 66.7 & 71.7 & 72.7 & 72.9 & 71.1 & 75.5 & 63.5 & 69.2 \\
\textbf{w. gold answer RL} & 56.8 & 65.6 & 60.0 & 62.8 & 66.4 & 71.2 & 73.6 & 75.6 & 67.2 & 79.6 & 60.8 & 67.9 & 59.5 & 66.8 & 67.3 & 72.7 & 72.3 & 73.6 & 75.4 & 75.4 & 75.3 & 79.0 & 64.5 & 71.7 \\
\textbf{Infer w. nllb pivot (PASMR)} & 55.2 & 52.8 & 60.8 & 58.0 & 62.4 & 68.0 & 71.2 & 76.4 & 70.8 & 80.4 & 56.3 & 65.6 & 57.5 & 63.3 & 61.5 & 69.3 & 68.1 & 68.9 & 71.4 & 74.8 & 71.2 & 76.5 & 60.8 & 68.2 \\
\textbf{Infer w. nllb pivot (PASMR wo. SRL)} & 52.4 & 51.2 & 52.0 & 52.0 & 60.0 & 62.0 & 66.0 & 70.4 & 66.8 & 76.4 & 51.9 & 60.9 & 56.9 & 60.3 & 60.0 & 67.9 & 67.1 & 65.7 & 70.9 & 71.3 & 68.0 & 75.3 & 59.1 & 66.3 \\
\textbf{Infer w. gold pivot} & 70.8 & 74.8 & 69.6 & 76.0 & 73.2 & 75.2 & 78.8 & 79.2 & 75.6 & 80.4 & 71.1 & 75.4 & 69.6 & 72.2 & 68.2 & 74.5 & 74.0 & 74.9 & 75.4 & 74.8 & 75.1 & 76.5 & 70.0 & 73.5 \\
\textbf{Infer w. gold pivot (PASMR wo. SRL)} & 66.8 & 72.0 & 65.6 & 70.4 & 74.8 & 71.2 & 73.2 & 74.4 & 71.2 & 76.4 & 68.1 & 71.6 & 67.9 & 70.7 & 65.2 & 73.3 & 70.6 & 71.8 & 73.4 & 74.4 & 72.9 & 75.8 & 67.9 & 71.6 \\

\midrule

\textbf{PASMR Deepseek-math-7b-instruct} & 60.8 & 69.6 & 49.2 & 69.2 & 76.4 & 77.2 & 78.8 & 79.6 & 76.4 & 86.4 & 59.9 & 72.4 & 68.2 & 75.1 & 69.7 & 80.9 & 82.4 & 82.0 & 81.2 & 83.1 & 82.6 & 85.4 & 71.0 & 79.1 \\
\textbf{wo. SRL} & 59.6 & 67.6 & 44.8 & 67.2 & 71.6 & 75.6 & 77.2 & 79.6 & 75.6 & 86.0 & 57.3 & 70.5 & 64.7 & 74.5 & 67.7 & 77.8 & 81.4 & 81.0 & 79.4 & 82.6 & 81.6 & 85.4 & 69.0 & 77.6 \\
\textbf{w. gold answer RL} & 61.2 & 69.2 & 50.0 & 72.0 & 76.0 & 80.0 & 79.2 & 80.8 & 75.2 & 86.0 & 60.1 & 73.0 & 65.5 & 76.6 & 66.5 & 78.7 & 82.1 & 82.3 & 82.5 & 83.7 & 82.9 & 86.7 & 69.5 & 78.8 \\
\textbf{Infer w. nllb pivot (PASMR)} & 64.0 & 59.6 & 61.6 & 68.0 & 68.0 & 76.0 & 78.4 & 80.0 & 73.2 & 86.0 & 61.7 & 71.5 & 65.6 & 68.8 & 70.1 & 78.1 & 78.6 & 80.0 & 80.4 & 82.9 & 80.5 & 85.4 & 68.2 & 77.0 \\
\textbf{Infer w. nllb pivot (PASMR wo. SRL)} & 60.4 & 62.0 & 59.6 & 65.2 & 69.2 & 75.2 & 77.6 & 80.0 & 69.6 & 85.6 & 60.7 & 70.4 & 63.7 & 68.0 & 69.3 & 76.2 & 78.5 & 78.6 & 79.3 & 82.0 & 80.3 & 84.8 & 67.0 & 76.1 \\
\textbf{Infer w. gold pivot} & 77.2 & 80.0 & 73.2 & 80.4 & 83.6 & 80.0 & 82.4 & 86.0 & 86.4 & 84.4 & 76.8 & 81.4 & 77.5 & 81.6 & 76.4 & 81.0 & 84.2 & 83.0 & 81.7 & 83.9 & 82.2 & 85.5 & 78.5 & 81.7 \\
\textbf{Infer w. gold pivot (PASMR wo. SRL)} & 72.4 & 81.2 & 73.2 & 78.0 & 79.2 & 83.6 & 83.2 & 83.2 & 84.8 & 85.2 & 75.6 & 80.4 & 74.7 & 79.0 & 74.9 & 77.9 & 83.0 & 82.1 & 81.8 & 82.6 & 81.4 & 85.5 & 76.2 & 80.3 \\

\bottomrule
\end{tabular}
}
\caption{Ablation Results. wo. SRL: no SRL stage; Infer w. nllb/gold pivot: infer with pivot from NLLB model or gold translations.}
\label{Ablation:MGSM&MSVAMP}
\end{table*}

\subsection{Main Results}

\textbf{PASMR Consistently Enhances Multilingual Reasoning.} 
As shown in Table~\ref{table:merged_results}, PASMR consistently improves performance across models and datasets, with larger gains on low-resource languages. For example, on MGSM, Mistral-7B-Instruct improves by 25.5\% (27 → 52.5) and Llama-3-8B-Instruct by 9.1\% (57.8 → 66.9). On low-resource languages, improvements of each model reach 32.8\% (11.6 → 44.4), 13.5\% (45.3 → 58.8), and 20.7\% (39.2 → 59.9). Compared to non-training baselines, PASMR also yields more stable gains, while pipeline methods show high variance (e.g., drops on DeepseekMath).

\begin{figure}[t]
\centering 
\includegraphics[scale=0.22]{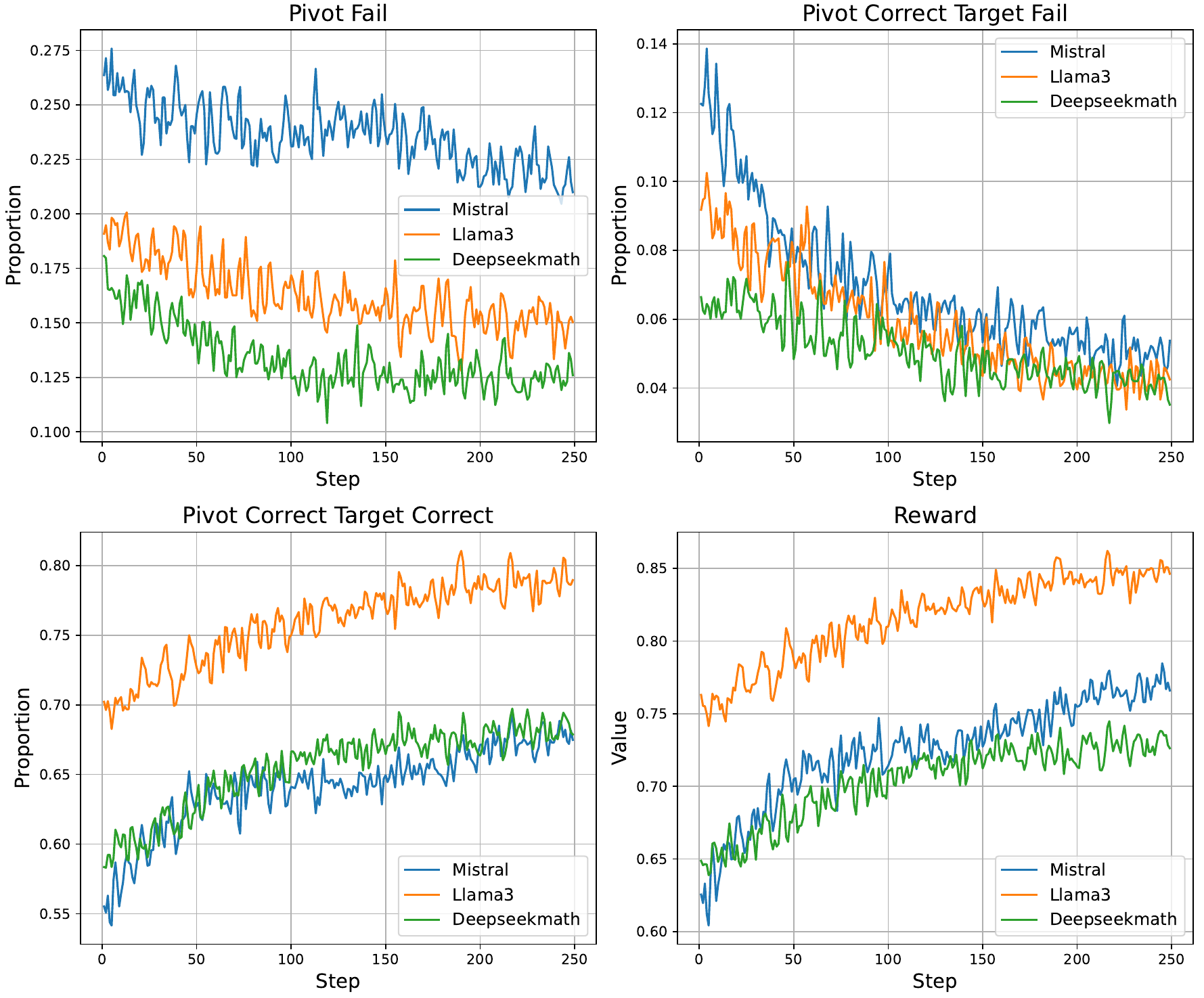} 
\caption{Metrics for RL training in SRL. Pivot Fail denotes the proportion of incorrect answers in the Pivot language. Pivot Correct Target Correct indicates cases where both Pivot and target answers are correct. Pivot Correct Target Fail refers to correct Pivot answers but incorrect target answers. Reward represents the average reward during training.} 
\label{trainlog:pivot} 
\end{figure}

\textbf{Pivot Language Alignment Boosts Reasoning.} 
Training-based baselines generally achieve greater improvements in multilingual reasoning compared to non-training approaches, while PASMR delivers additional gains beyond these methods. For instance, on MSVAMP, Deepseek-math-7B-Instruct rises from 70 to 79.1 with PASMR, compared to 75.2 with Gold Answer RL. This highlights pivot alignment as the key driver of performance. Moreover, Figure~\ref{trainlog:pivot} shows that increasing reward values lead to higher accuracy in pivot-aligned answers and fewer errors, suggesting that pivot alignment strengthens reasoning in the target language.


\textbf{Robustness and Generalization.} PASMR shows strong robustness and consistent generalization across both in-domain and out-of-distribution (OOD) data. During the SRL stage, PASMR maintains performance on OOD data, e.g., Llama with PASMR achieves 66.9/71.3 on MGSM and MSVAMP, while with OOD data it reaches 67.0/74.0, demonstrating resilience to data shifts. In contrast, MSFT improves in-domain accuracy but often drops on out-of-domain sets (MGSM 57.8→59.0, MSVAMP 69.5→64.1). PASMR consistently boosts performance across datasets, supporting continuous learning from new, unlabeled data.

\subsection{Ablation Study \& Discussion}
\textbf{Effect of SRL on Training and Inference.} 
To investigate the role of SRL and pivot-based rewards, we conduct ablation studies, with results shown in Table~\ref{Ablation:MGSM&MSVAMP}. On the training side, removing SRL (wo. SRL) consistently reduces performance, highlighting the value of pivot-language reasoning alignment. Compared to using ground-truth answers as rewards (w. gold answer RL), our self-generated signals achieve similar or better results, showing pivot-based rewards are a simpler yet effective solution.

On the inference side, we replace the pivot language with English translations from the professional translation model (Infer w. nllb pivot)\footnote{https://huggingface.com/facebook/nllb-200-3.3B}
 to avoid bias from self-generated translations. Removing SRL again degrades performance, underscoring its role in multilingual reasoning. In addition, professional translations slightly underperform self-generated ones, suggesting our method better exploits the model’s inherent multilingual alignment.

\begin{figure}[t]
\centering 
\includegraphics[width=0.92\linewidth]{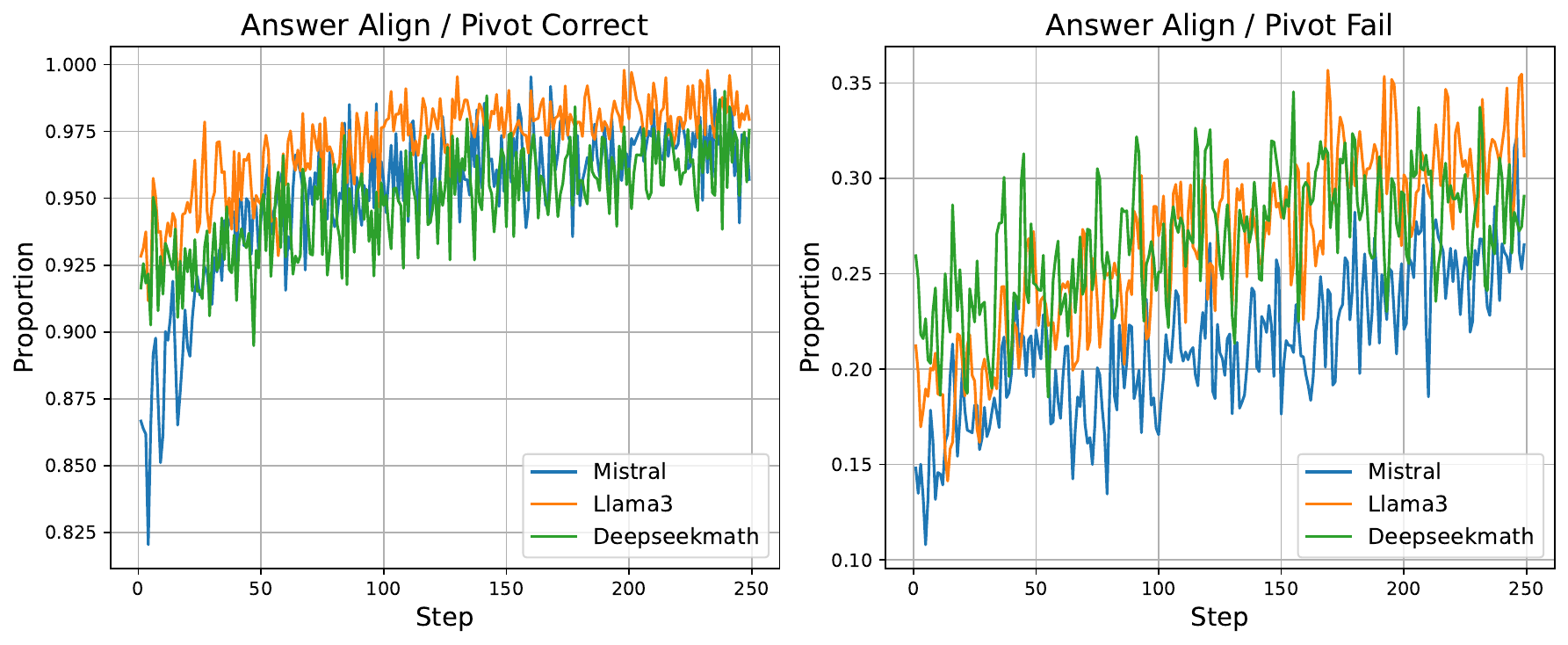} 
\caption{Alignment Metrics. Answer Align / Pivot Correct — target matches pivot on correct answers; Answer Align / Pivot Fail — target matches pivot on incorrect answers.} 
\label{trainlog:pivot quality} 
\end{figure}

\textbf{Pivot Language Quality Matters.}
In multilingual reasoning tasks, answer consistency between the target and pivot language is largely determined by the pivot’s correctness. As shown in Figure~\ref{trainlog:pivot quality}, consistency reaches 82–98\% when the pivot is correct but falls to 12–35\% when it is wrong, highlighting the pivot’s critical role. Pivot correctness depends on (1) its quality (translation accuracy and guidance effectiveness) and (2) task difficulty relative to the model’s reasoning ability. To disentangle these factors, we replace the pivot language with the corresponding English version as the "gold pivot",  which significantly improves performance (Infer w. gold pivot in Table~\ref{Ablation:MGSM&MSVAMP}), especially for low-resource languages, confirming the importance of pivot quality in reasoning alignment.


\textbf{Variation in Performance Gains Across Languages and Models.}  
We observe two patterns:  
(1) Languages with much lower baseline performance than English, such as Swahili and Bengali, show the largest gains. This mainly comes from PAM, which improves consistency in mapping problems into the pivot language.  
(2) Models with weaker reasoning, such as Mistral, also benefit more because training strengthens their core ability. In contrast, stronger models (e.g., LLaMA-3, DeepSeek) already perform well, so their gains are mostly in multilingual consistency.

\begin{figure}[t]
\centering 
\includegraphics[scale=0.23]{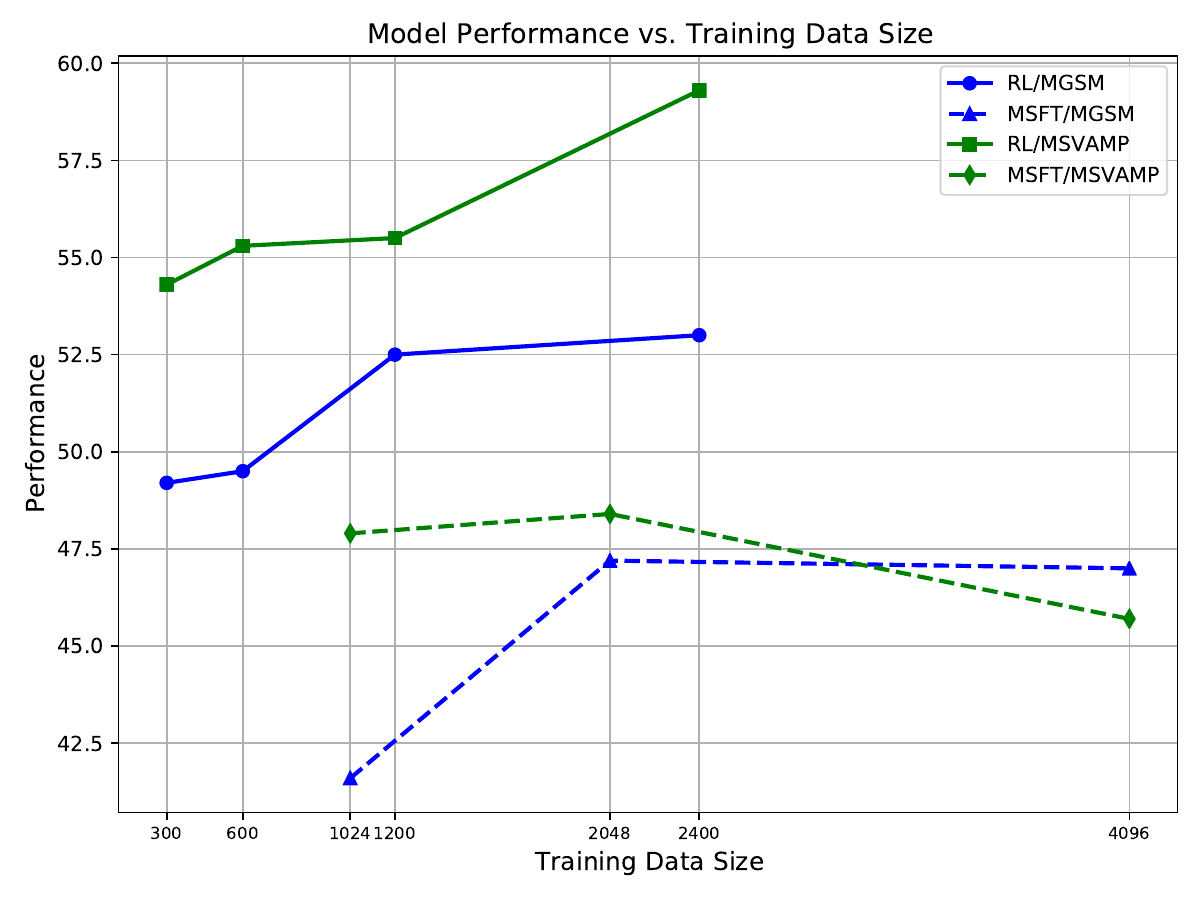} 
\caption{Performance of Mistral models with SRL and MSFT as training data size per language increases.} 
\label{train_size} 
\end{figure}

\textbf{Impact of Training Data Size on RL and SFT-Based Methods.}
To study the effect of training data size, we varies training data size per language (Figure~\ref{train_size}). SRL performance steadily improves with more data on both in-domain (MGSM) and out-of-domain (MSVAMP) sets. In contrast, MSFT improves only initially and then saturates or even declines (e.g., 48.4 → 45.7 on MSVAMP), suggesting stronger memorization and weaker generalization than SRL.

\section{Related Work \& Conclusion}
LLMs’ unbalanced multilingual reasoning abilities largely stem from uneven data distribution during training. A common solution is post-training with multilingual-augmented data \cite{chai2024xcot, zhang2024enhancing}. Such data augmentation helps build a multilingual-aligned representation space, thereby enhancing cross-lingual generalization.

Another line of work leverages high-resource languages. In the training-free setting, models first reason in the primary language and then translate the results into the target language \cite{etxaniz2023multilingual, huang2023not, han2025loire, zhang2023don}. These pipeline-based approaches require prompt tuning and multiple inference steps to obtain final answers. Training-based methods instead construct preference pairs from cross-lingual answer discrepancies and apply preference learning to improve reasoning. For example, MAPO \cite{she2024mapo} employs a translation model to evaluate the quality of multilingual reasoning answers, thereby reducing the likelihood of generating lower-scoring outputs. Similarly, Yang et al. \cite{yang2024language} ranks reasoning answers in both primary and non-primary languages, as well as their translations, leveraging DPO to optimize the model's multilingual reasoning choices. Such methods often rely on external translation models or supervision singals.

In contrast, our approach introduces the PASMR framework, which aims to address multilingual reasoning challenges by using the primary language as a pivot and leveraging self-generated reward signals, without requiring gold-standard answers or external models. Experimental results demonstrate that PASMR significantly improves multilingual math reasoning performance, particularly for low-resource languages, while maintaining robust generalization and consistency. These findings provide promising directions and insights for advancing multilingual reasoning in LLMs.

\section{acknowledgements}
This work was supported by research project of Nanjing University-China Mobile Joint Institute (NJ20250038) and JIUTIAN Research.

\label{sec:refs}



\bibliographystyle{IEEEbib}
\bibliography{strings,main}

\end{document}